\documentclass[runningheads]{llncs}
\usepackage[T1]{fontenc}
\usepackage{graphicx}
\usepackage[mathscr]{eucal}
\usepackage{amsmath}
\usepackage{amssymb}
\usepackage{multirow}
\usepackage[table,xcdraw]{xcolor}

\usepackage{bm}
\usepackage{ulem}
\usepackage{xcolor}
\usepackage{algorithm}
\usepackage{algpseudocode}
\usepackage{amsmath}
\usepackage{footnote}

\usepackage{booktabs}
\usepackage{marvosym}

\makeatletter
\def\thanks#1{\protected@xdef\@thanks{\@thanks
        \protect\footnotetext{#1}}}
\makeatother

\begin{document}
\NewDocumentCommand\emojionee{}{\scalerel{\includegraphics[width=1.2em]{figs/logo.png}{}}}
\NewDocumentCommand\emojione{}{%
  \includegraphics[width=1.1em]{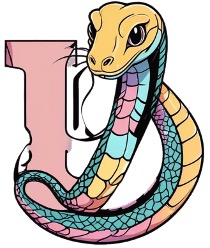}{}%
}

\title{\emojione{}LKM-UNet: Large Kernel Vision Mamba UNet \\ for Medical Image Segmentation}
\titlerunning{Large Kernel Vision Mamba UNet for Medical Image Segmentation}
% \textsuperscript{(\Letter)}
% \renewcommand{\thefootnote}{\Letter}

\author{Jinhong Wang\inst{1,2,5} \and
Jintai Chen\inst{3}\textsuperscript{(\Letter)} \and
Danny Chen\inst{4} \and
Jian Wu\inst{5,6}
\thanks{\Letter : Corresponding Author.}
}

\authorrunning{J. Wang et al.}

\institute{College of Computer Science and Technology, Zhejiang University, China\\ \and 
Institute of Wenzhou, Zhejiang University, China \\ \and
Computer Science Department, University of Illinois Urbana-Champaign, USA\\ \and
Department of Computer Science and Engineering, University of Notre Dame, USA \\ \and
State Key Laboratory of Transvascular Implantation Devices of The Second Affiliated Hospital, Zhejiang University School of Medicine, China\\ \and
School of Public Health Zhejiang University, China\\ 
\email{wangjinhong@zju.edu.cn}, \email{jtchen721@gmail.com}, \email{dchen@nd.edu}, \email{wujian2000@zju.edu.cn}}
\maketitle              % typeset the header of the contribution
\begin{abstract}

In clinical practice, medical image segmentation provides useful information on the contours and dimensions of target organs or tissues, facilitating improved diagnosis, analysis, and treatment.
In the past few years, convolutional neural networks (CNNs) and Transformers have dominated this area, but they still suffer from either limited receptive fields or costly long-range modeling. Mamba, a State Space Sequence Model (SSM), recently emerged as a promising paradigm for long-range dependency modeling with linear complexity. In this paper, we introduce a {\bf L}arge {\bf K}ernel vision {\bf M}amba {\bf U}-shape {\bf Net}work, or \texttt{LKM-UNet}, for medical image segmentation. A distinguishing feature of our \texttt{LKM-UNet} is its utilization of large Mamba kernels, excelling in locally spatial modeling compared to small kernel-based CNNs and Transformers, while maintaining superior efficiency in global modeling compared to self-attention with quadratic complexity. Additionally, we design a novel hierarchical and bidirectional Mamba block to further enhance Mamba's global and neighborhood spatial modeling capability for vision inputs. Comprehensive experiments demonstrate the feasibility and the effectiveness of using large-size Mamba kernels to achieve large receptive fields. Codes are available at \url{https://github.com/wjh892521292/LKM-UNet}.

\end{abstract}
\section{Introduction}
Efficiently segmenting biomedical objects of interest (\textit{e.g.}, lesions) in large-size 2D/3D images significantly enhances downstream clinical practice and biomedical research. Currently, automatic segmentation models, leveraging popular deep learning backbones such as convolutional neural networks (CNNs) and Transformers~\cite{vaswani2017attention}, have reduced manual annotation requirements but may incur considerable computational costs or overlook specific details. CNN-based models (e.g., UNet~\cite{ronneberger2015u}) typically seek to extract global patterns by hierarchically stacking small kernels, excelling in pixel-level feature extraction but being ineffective in learning long-range dependencies due to their limited receptive fields~\cite{luo2016understanding}. Though recent investigations~\cite{repvgg} have shown the effectiveness of large convolution kernels, it often requires specific optimization strategies and complicated inference-time model reformulation. In contrast, Transformer-based algorithms offer powerful long-range modeling but sacrifice pixel-level spatial
modeling~\cite{wu2023d,wu2021cvt}. Further, a key component, the self-attention module, incurs quadratic complexity and cannot handle excessive tokens~\cite{hua2022transformer}, resulting in the need to pack pixels into windows and hence sacrificing resolution information. Especially, many studies have shown that Transformers achieve the best trade-off at a $7 \times 7$ window size as a smaller window causes more computation and a larger window causes a significant drop in performance~\cite{liu2021swin,chu2021twins}.
Later studies have investigated CNN-Transformer hybrid models or approaches for intra-patch locality modeling~\cite{wu2021cvt,yuan2021incorporating}. However, due to the typically larger sizes of medical images compared to natural images, how to reduce the interaction complexity between global patches and how to enlarge the receptive field of local spatial modeling are still open.

Recently, structured state space sequence models (SSMs)~\cite{gu2021efficiently} (e.g., Mamba~\cite{gu2023mamba}) have emerged as a powerful long-sequence modeling approach with linear complexity in terms of input size, shedding light on efficient modeling of both local and global dependencies. In contrast to the conventional self-attention approach, SSMs enable each element in a 1D array (e.g., a text sequence) to interact with any of the previously scanned samples through a compressed hidden state, effectively reducing the quadratic complexity to linear. SSMs were devised to address natural language processing (NLP) tasks~\cite{fu2022hungry}, but also show effectiveness in computer vision~\cite{zhu2024vision}.
For medical image segmentation, 
%our cocurrent work, 
for example, U-Mamba~\cite{gu2023mamba} and SegMamba~\cite{xing2024segmamba} introduced SSM-CNN hybrid models that directly apply Mamba to efficiently model long-range dependencies at the pixel-level. Although effective, the potential and deficiency of Mamba are still not yet fully explored and resolved, in three aspects. {\bf First}, benefiting from its linear complexity, Mamba possesses more flexibility in space allocation. That is, unlike the stereotypes of previous methods based on small convolution kernels or size-constrained self-attention windows, Mamba is promising in endowing the model with an ability in large receptive field spatial modeling, which has been neglected in the known Mamba-based approaches. {\bf Second}, Mamba is a unidirectional sequence modeling method that lacks positional awareness and focuses more on posterior tokens. {\bf Third}, the original Mamba was proposed for 1D sequence modeling, which is not suitable for computer vision tasks that require spatial-aware understanding. Mainly due to local adjacent spatial pixels becoming discontinuous, Mamba can suffer forgetting problems and inefficient local modeling capabilities.

In this paper, we propose a {\bf L}arge {\bf K}ernel {\bf M}amba {\bf U}-shape {\bf Net}work (\texttt{LKM} \texttt{-UNet}) for 2D and 3D medical image segmentation.   \texttt{LKM-UNet} utilizes the powerful sequence modeling capabilities and linear complexity of Mamba to implement large receptive fields by assigning large kernels (windows) to SSM modules. Further, we design a novel hierarchical and bidirectional large kernel Mamba block (LM block) to enhance the representation modeling capability of SSMs. %Specifically, 
The bidirectional Mamba design is used for location-aware sequence modeling, reducing the weight impact of the input order. The hierarchical Mamba design is composed of two types of SSM operations: (i)
pixel-level SSM (PiM) and (ii) patch-level SSM (PaM). PiM captures the neighborhood and local-scope pixel information by large kernel SSM and avoids the forgetting problem that occurs in SSMs due to discontinuity of adjacent patches after tokenization. PaM deals with long-range dependency modeling and global patch interaction. 

Our main contributions are as follows. (1) We propose a Large Kernel Mamba UNet (\texttt{LKM-UNet}) for 2D/3D medical image segmentation. (2) We assign kernels of large receptive fields to SSM layers to enable the model to possess the capability of large spatial modeling. (3) We design a bidirectional Mamba for location-aware sequence modeling. (4) We propose a novel hierarchical Mamba module composed of pixel-level SSM (PiM) and patch-level SSM (PaM), enhancing local-neighborhood pixel-level and long-range global patch-level modeling.

\section{Preliminaries: SSM Models} 
SSM-based models, namely the structured state space sequence models (S4) and Mamba~\cite{gu2023mamba}, all evolved from the continuous system that maps a
1-dimensional function or sequence $x(t) \rightarrow y(t) \in \mathcal{R}$ through a hidden state $h(t) \in \mathcal{R}^{N}$. This process can be  represented as
the following linear ordinary differential equation:
\vspace{-0.3em}
\begin{equation}
\begin{split}
     h^{\prime}(t) &= \mathbf{A}h(t) + \mathbf{B}x(t), \\
    y(t) &= \mathbf{C}h(t),
\end{split}
\label{ode}
\end{equation}
\vskip -0.1em
\noindent
where $\mathbf{A} \in \mathcal{R}^{N \times N}$ is a state matrix and $\mathbf{B},\mathbf{C} \in \mathcal{R}^{N}$ are projection parameters.

S4 and Mamba are discrete versions of the aforementioned continuous system, which include a timescale parameter $\mathbf{\Delta}$ to
transform the continuous parameters $\mathbf{A}, \mathbf{B}$ to discrete parameters $\overline{\mathbf{A}}, \overline{\mathbf{B}}$. Typically, the zero-order hold (ZOH) is employed as the discretization rule and can be defined as follows:
\vspace{-0.3em}
\begin{equation}
\begin{split}
    \overline{\mathbf{A}} &= \text{exp}(\Delta \mathbf{A}), \\
    \overline{\mathbf{B}} &= (\mathbf{\Delta} \mathbf{A})^{-1}(\text{exp}(\mathbf{\Delta} \mathbf{A}) - \mathbf{I}) \cdot \Delta \mathbf{B}.
\end{split}
\label{zoh}
\end{equation}
\vskip -0.1em
\noindent
After the discretization, the discretized version of Eq.~(\ref{ode}) can be defined as :
\vspace{-0.25em}
\begin{equation}
\begin{split}
     h^{\prime}(t) &= \overline{\mathbf{A}}h(t) + \overline{\mathbf{B}}x(t), \\
    y(t) &= \mathbf{C}h(t).
\end{split}
\label{dis_ode}
\end{equation}
\vskip -0.1em
\noindent
Then the output is computed through a global convolution, defined as:
\vspace{-0.3em}
\begin{equation}
\begin{split}
    \overline{\mathbf{K}} &= (\mathbf{C}\overline{\mathbf{B}}, \mathbf{C}\overline{\mathbf{AB}}, \mathbf{C}\overline{\mathbf{A}}^{L-1}\overline{\mathbf{B}}), \\
    y &= x * \overline{\mathbf{K}},
\end{split}
\label{out}
\end{equation}
\vskip -0.1em
\noindent
where $L$ is the length of the input sequence $x$ and $\overline{\mathbf{K}} \in \mathcal{R}^{L}$ is a structured convolutional kernel.

\section{Method}
In this section, we first introduce the overall architecture of \texttt{LKM-UNet}. Subsequently, we elaborate the core component, the LM block. 

% {\bf \quad} 

\begin{figure}[t]
\centering
\includegraphics[width=0.98\textwidth]{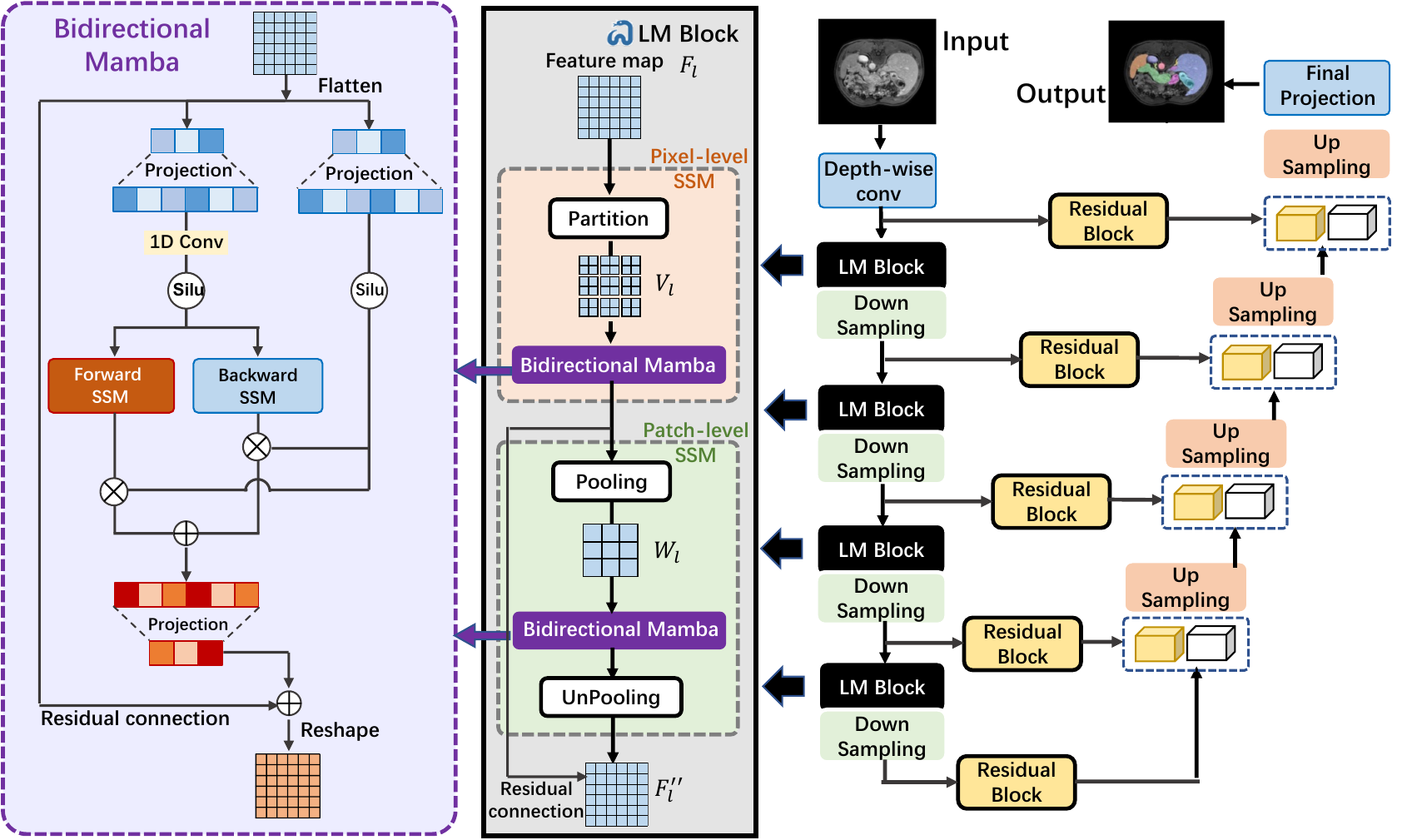}
\caption{An overview of our proposed \texttt{LKM-UNet}.}
\label{overview}
\vskip -1 em
\end{figure}

\subsection{LKM-UNet}
An overview 
%pipeline 
of \texttt{LKM-UNet} is given in Fig.~\ref{overview}. Specifically, except for the common UNet composition of a depth-wise convolution, an encoder with downsampling layers, a decoder with upsampling layers, and skip-connection, \texttt{LKM-UNet} improves the structure of UNet with proposed large kernel Mamba (LM) blocks inserted into the encoder. Given a 3D input image with a resolution of $C \times D \times H \times W$, the depth-wise convolution first encodes the input into a feature map $\mathbf{F}_{0} \in \mathcal{R}^{48 \times \frac{D}{2} \times \frac{H}{2} \times \frac{W}{2}}$. Then the feature map  $\mathbf{F}_{0}$ is fed into each LM block and the corresponding down-sampling layers, and multi-scale feature maps are obtained. An LM block contains two Mamba modules: pixel-level SSM (PiM) and patch-level SSM (PaM). For the $l^{th}$ layer, the process can be formulated as:
\begin{equation}
    \mathbf{F^{\prime}}_{l} = \text{PiM}(\mathbf{F}_l),  \quad
    \mathbf{F^{\prime\prime}}_{l} = \text{PaM}(\mathbf{F^{\prime}}_{l}),  \quad
    \mathbf{F}_{l+1} = \text{Down-sampling}(\mathbf{F^{\prime\prime}}_{l}),
    \label{LM}
\end{equation}
where PiM and PaM denote pixel-level SSM and patch-level SSM, respectively. Down-sampling denotes the down-sampling layers. After each stage, the produced feature map $\mathbf{F}_{l+1}$ is encoded to $(2C_{l}, \frac{D_{l}}{2}, \frac{H_{l}}{2},\frac{W_{l}}{2}$), where $C_{l}, (D_{l}, H_{l}, W_{l})$ represent the channel and resolution of feature map $\mathbf{F}_{l}$. As for the decoder part, we adopt a UNet decoder and residual block with skip connections for upsampling and predicting the final segmentation masks.

\subsection{LM block}
\begin{figure}[t]
\centering
\includegraphics[width=0.98\textwidth]{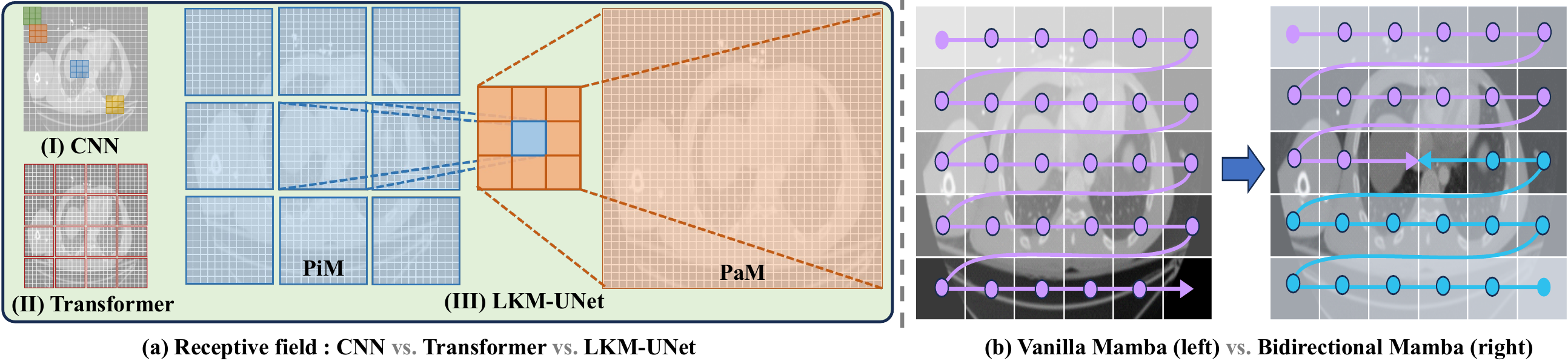}
\caption{(a) Respective field comparison among CNN, Transformer, and our proposed \texttt{LKM-UNet}. CNNs often use small kernels (like $3 \times 3$), and Transformers often use $7 \times 7$ sized kernels (windows). Our LKM-UNet can scale up kernel size to $40 \times 40$. (b) Scanning order comparison of vanilla Mamba vs. our proposed bidirectional Mamba.}
\label{pic1}
\vskip -1 em
\end{figure}

LM block is our core component used for further spatial modeling of feature maps of different scales at each stage. Different from the previous methods that use CNN for local pixel-level modeling and Transformer for long-range patch-level dependency modeling, an LM block can accomplish pixel-level and patch-level modeling simultaneously, benefiting from the linear complexity of Mamba. More critically, a lower complexity allows setting larger kernels (windows) to obtain bigger receptive fields, which will improve the efficiency of local modeling, as shown in Fig.~\ref{pic1}(a). Specifically, LM block is a hierarchical design that consists of pixel-level SSM (PiM) and patch-level SSM (PaM); the former is used for local-neighborhood pixel modeling and the latter is used for global long-range dependency modeling. 
Besides, each Mamba layer in an LM block is bidirectional, which is proposed for location-aware sequence modeling.

\noindent
{\bf Pixel-level SSM (PiM).}
Since Mamba is a continuous model, the discrete nature of input pixels can weaken the correlation modeling of locally adjacent pixels. Hence, we propose a pixel-level SSM to split the feature map into multiple large sub-kernels (sub-windows) and perform SSM operations on the sub-kernels. We first equally divide a whole feature map into non-overlapping sub-kernels for 2D or sub-cubes for 3D. Take 2D for example. Given an input of $H \times W$ resolution, we divide the feature map into sub-kernels of size $m \times n$ each ($m$ and $n$ can be up to 40). Without loss of generality, we assume that $H/m$ and $W/n$ are both integers. Then we have $\frac{HW}{mn}$ sub-kernels, as shown in Pixel-level SSM of Fig.~\ref{overview}. Under this scheme, when these sub-kernels are sent into a Mamba layer, the local adjacent pixels will be input continuously into SSM; thus, the relation between local-neighborhood pixels can be better modeled. Moreover, under the large kernel partition strategy, the receptive fields are enlarged and the model can obtain more details of the local pixels. However, the image is divided into non-overlapping sub-kernels. Hence, we need a mechanism for communication between different sub-kernels, for long-range dependency modeling.

\noindent
{\bf Patch-level SSM (PaM).}
We introduce a patch-level SSM layer to impart information among different sub-kernels. As shown in Patch-level SSM of Fig.~\ref{overview}, a feature map $\mathbf{F^{\prime}}_{l}$ of resolution $H \times W$ first passes through a pooling layer of size $m \times n$ to allow important information for each of the $\frac{HW}{mn}$
%$m \times n$
sub-kernels to be summarized into a single representative. Thus, we obtain aggregate maps $\mathbf{Z}_{l}$ with $\frac{HW}{mn}$ representatives, which are then used to communicate among the sub-kernels through Mamba for global-range dependency modeling. After the interaction in Mamba, we unpooling the aggregate maps back to the same size as the initial feature map $\mathbf{F^{\prime}}_{l}$, and apply a residual connection. The process of $\mathbf{F}^{\prime\prime}_l = \text{PaM}(\mathbf{F}^{\prime}_{l})$ in Eq.~(\ref{LM}) can be carried out as:
\begin{equation}
    \mathbf{W}_{l} = \text{Pooling}(F^{\prime}_{l}), \quad
    \mathbf{W}^{\prime}_{l} = \text{Bi-Mamba}(\mathbf{W}_{l}), \quad
    \mathbf{F^{\prime\prime}}_{l} = \text{Unpooling}(\mathbf{W}^{\prime}_{l}), 
    \label{PaM}
\end{equation}
where Pooling and Unpooling denote the pooling layer and unpooling layer, respectively. Bi-Mamba denotes the proposed bidirectional Mamba layer.

\noindent
{\bf Bidirectional Mamba (BiM).}
Different from the vanilla Mamba block which is based on forward-only scan direction SSM layers, each SSM layer (including PiM and PaM) in our LM block is bidirectional. Fig.~\ref{pic1}(b) shows the differences. In the original Mamba, as a continuous model, some information forgetting occurs on the elements entered earlier, and the latest elements that enter into Mamba will retain much more information. Thus, the original Mamba with a single scanning direction will focus more on the posterior patches, rather than the center areas of the feature maps often with more organs and lesions. To this end, we propose a bidirectional Mamba structure by performing both forward and backward scanning at the same time and superimposing the output results. The detailed structure is shown in the left part of Fig.~\ref{overview}. BiM has two advantages. First, the model can focus more on the informative patches in the center areas of the image likely with more organs and lesions, rather than the corner areas. Second, for each patch, the absolute position information and relative position information with other patches can be well modeled by the network.

\section{Experiments}

\subsection{Datasets}

We conduct experimental comparisons with state-of-the-art methods on two datasets for 2D and 3D segmentation tasks to validate the effectiveness and scalability of \texttt{LKM-UNet}.

\noindent
{\bf Abdomen CT.}
Abdomen CT is a publicly available 3D multi-organ segmentation dataset comprising 100 CT cases from the MICCAI 2022 FLARE Challenge~\cite{ma2023unleashing}, including 13 types of abdominal organs (liver, spleen, pancreas, right kidney, left kidney, stomach, gallbladder, esophagus, aorta, inferior vena cava, right adrenal gland, left adrenal gland, and duodenum). The size of a 3D CT image is 40 $\times$ 224 $\times$ 192. 50 cases from the MSD Pancreas dataset with annotations from AbdomenCT-1K are used for training, and another 50 cases from different medical centers~\cite{clark2013cancer} are used for testing.

\noindent
{\bf Abdomen MR.}
Abdomen MR is a publicly available 2D segmentation dataset comprising 110 MRI cases from the MICCAI 2022 AMOS Challenge~\cite{ji2022amos}, including 13 types of abdominal organs (the same as the Abdomen CT dataset). The size of a 2D MRI image is 320 $\times$ 320. Following the previous work~\cite{ma2024u,xing2024segmamba}, 60 annotated cases are used for training, and another 50 cases are used for testing.

\renewcommand{\cmidrulesep}{0.5mm} %定义两条相邻
\setlength{\aboverulesep}{0.5mm} %在线条[不包括
\setlength{\belowrulesep}{0.5mm} %在线条[不包括

\begin{table}[t]
\centering
\caption{Quantitative segmentation results on the 3D Abdomen CT dataset and 2D Abdomem MR dataset.}
\label{tab:my-table}
\tabcolsep=0.2cm
\renewcommand\arraystretch{1.2}
\resizebox{0.9\textwidth}{!}{%
\begin{tabular}{c|lccl|lccl}
\toprule
                         &  & \multicolumn{2}{l}{\ \ 3D Abdomen CT \quad } &  &  & \multicolumn{2}{l}{\ \ 2D Abdomen MR} &  \\ 
                          \cline{3-4}\cline{7-8}
\multirow{-2}{*}{Method} &  & \quad DSC  $\uparrow$            & \quad NSD $\uparrow$ \quad           &  &  & \quad DSC $\uparrow$          & \quad NSD $\uparrow$ \quad               &  \\ \midrule
SegResNet~\cite{myronenko20193d}                &  & 79.27               & 82.57               &  &  & 73.17             & 80.34                   &  \\
nnUNet~\cite{isensee2021nnu}                 &  & 86.15               & 89.72               &  &  & 74.50             & 81.53                   &  \\
% MedNeXt                  &  & 1               & 1               &  &  & 1             & 1        &  
\midrule
UNTER~\cite{hatamizadeh2022unetr}                    &  & 68.24               & 70.04               &  &  & 57.47             & 63.09                   &  \\
SwinUNTER~\cite{hatamizadeh2021swin} &
   &
75.94 &
76.63 &
   &
   &
  70.28 &
  76.69 &
   \\
nnFormer~\cite{zhou2023nnformer} &
   &
  78.34 &
  81.45 &
   &
   &
  72.79 &
  79.63 &
   \\ \midrule
  U-Mamba (concurrent)~\cite{ma2024u} &
   &
  86.38 &
  89.80 &
   &
   &
  76.25 &
  83.27 & 
  \\ \midrule
\cellcolor[HTML]{E2E0E0}\textbf{\texttt{LKM-UNet}(Ours)} & \cellcolor[HTML]{E2E0E0}
   &
  \cellcolor[HTML]{E2E0E0}\textbf{86.82} &
  \cellcolor[HTML]{E2E0E0}\textbf{90.02} &
  \cellcolor[HTML]{E2E0E0} &
  \cellcolor[HTML]{E2E0E0} &
  \cellcolor[HTML]{E2E0E0}\textbf{77.35} &
  \cellcolor[HTML]{E2E0E0}\textbf{83.80} & \cellcolor[HTML]{E2E0E0}
   \\ \bottomrule
\end{tabular}%
}
\vskip -1 em
\end{table}

\subsection{Implementation Setup}
Our \texttt{LKM-UNet} is implemented on PyTorch 1.9.0 based on the nnU-Net framework. All the experiments are conducted on an NVIDIA GeForce RTX 3090 GPU. The batch size in training is 2 for the 3D dataset (Abdomen CT) and 24 for the 2D dataset (Abdomen MR). The Adam~\cite{kingma2014adam} optimizer with momentum = 0.99 is used. The initial learning rate is 0.01 with a weight decay of 3e-5. The maximum training epoch number is 1000. For the Abdomen CT dataset, the stage is 6 but the dimensions are not consistent; thus we set the rectangle kernel size with three dimensions to [20, 28, 24], [20, 28, 24], [10, 14, 12], [10, 14, 12], [5, 7, 6], and [5, 7, 6] for each stage. For the Abdomen MR dataset, the stage is 7 and the kernel size is 40, 20, 20, 10, 10, 5, and 5 for each stage.

\subsection{Overall Performances}
The baseline models include three types of representative networks: CNN-based networks (nnU-Net~\cite{isensee2021nnu} and SegResNet~\cite{myronenko20193d}), Transformer-based networks (UNETR~\cite{hatamizadeh2022unetr}, SwinUNETR~\cite{hatamizadeh2021swin}, nnFormer~\cite{zhou2023nnformer}), and the latest Mamba-based network (U-Mamba~\cite{ma2024u}). For fair comparison, we also implement all the models in the nnU-Net framework, and use the default image pre-processing. Table~\ref{tab:my-table} presents the results. Compared to both CNN-based and Transformer-based segmentation methods, our proposed \texttt{LKM-UNet} achieves improved performances in
both DSC and NSD, which indicates that the global modeling capabilities of Mamba are critical to medical image segmentation. Note that, compared to U-Mamba which simply applies Mamba as a global modeling adapter, \texttt{LKM-UNet} exhibits %remarkable 
improvements over U-Mamba, validating the effectiveness of our bidirectional and hierarchical Mamba designs. These results also demonstrate the potential of Mamba in global and local feature modeling with larger receptive fields.

% \usepackage{multirow}
\begin{table}[t]
\centering
\caption{Performances of \texttt{LKM-UNet} in three different kernel size settings. The kernel size sequence indicates the kernel size in each stage (the total number of stages is 7).}
\label{tab3}
\tabcolsep=0.2cm
\resizebox{0.82\textwidth}{!}{%
\begin{tabular}{c|c|c|c}
\toprule
\multirow{2}{*}{Kernel size} &
  \multirow{2}{*}{[10, 5, 5, 5, 5, 5, 5]} &
  \multirow{2}{*}{[20, 10, 10, 10, 5, 5, 5]} &
  \multirow{2}{*}{[40, 20, 20, 10, 10, 5, 5]} \\
    &       &     &       \\ \midrule
DSC & 75.89 & 76.45  & {\bf 77.35} \\ \midrule
NSD & 82.26   & 82.78  & {\bf 83.80} \\ \bottomrule
\end{tabular}
}
\vskip -0.5 em
\end{table}

% \usepackage{multirow}
\begin{table}[t]
\centering
\caption{Performances of \texttt{LKM-UNet} with different sub-modules. PiM = Pixel-level SSM. PaM = Patch-level SSM. BiM = Bidirectional Mamba.}
\label{tab2}
\tabcolsep=0.2cm
\resizebox{\textwidth}{!}{%
\begin{tabular}{c|c|c|c|cl|c|c|c}
\toprule
\multirow{2}{*}{Method} &
  \multirow{2}{*}{Baseline} &
  \multirow{2}{*}{Only PiM} &
  \multirow{2}{*}{Only PaM} &
  \multicolumn{2}{c|}{\multirow{2}{*}{PiM + BiM}} &
  \multirow{2}{*}{PaM + BiM} &
  \multirow{2}{*}{PiM + PaM} &
  \multirow{2}{*}{PiM + PaM + BiM} \\
    &       &   &   & \multicolumn{2}{c|}{}     &     &  &   \\ \midrule
DSC & 74.50 & 76.82 & 76.22 & \multicolumn{2}{c|}{76.90} & 76.73 & 77.10 & {\bf 77.35} \\ \midrule
NSD & 81.53     & 83.05 & 82.59 & \multicolumn{2}{c|}{83.31} & 82.94 & 83.54 & {\bf 83.80} \\ \bottomrule
\end{tabular}
}
\vskip -2 em
\end{table}

\subsection{Is the Kernel Size of \texttt{LKM-UNet} Important?}
To validate Mamba's large spatial modeling potential, we explore \texttt{LKM-UNet}'s performance in different kernel size settings. Table~\ref{tab3} shows the results on the Abdomen MR dataset. Comparing the performances of the three kernel-size settings, one can see that \texttt{LKM-UNet} with larger kernel sizes achieves better performances. This indicates that large receptive fields are critical for medical image segmentation which can be achieved with Mamba due to its linear complexity. 

\subsection{Ablation Study}
We conduct ablation experiments on the Abdomen MR dataset to validate the effect of each key component in our \texttt{LKM-UNet}, shown in Table~\ref{tab2}. Both PiM and PaM provide improvements for \texttt{LKM-UNet} over the baseline model, validating the superiority of PiM and PaM in local pixel-level modeling and global modeling, respectively. Notably, the model with PiM gains more improvements, suggesting that enlarging the receptive field of local modeling is a key to improving model performance. After introducing BiM, the performance of \texttt{LKM-UNet} further improves, which shows the importance of bidirectional Mamba for location-aware modeling. Finally, \texttt{LKM-UNet} with all the components achieves the best performance, further demonstrating our method's effectiveness and its components.

\subsection{Effective Receptive Field Visualization}
\begin{figure}[t]
\centering
\includegraphics[width=0.6\textwidth]{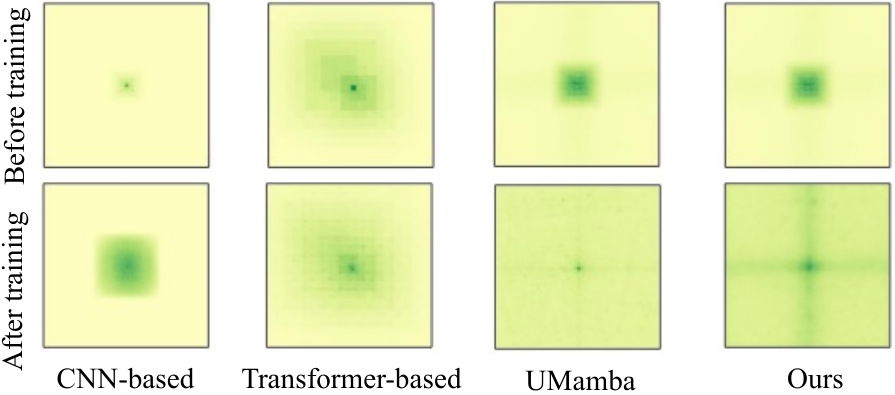}
\caption{Effective respective field visualization among CNN, Transformer, U-Mamba and our proposed \texttt{LKM-UNet}.}
\label{pic3}
\vskip -1 em
\end{figure}

To show more details of receptive field, we exhibit the effective receptive field (ERF)~\cite{luo2016understanding} of other methods and LKM-UNet in Fig.~\ref{pic3}. CNN-based methods focus more on local feature extraction, while Transformer-based methods have a wider range of ERF. Although U-Mamba utilize Mamba to obtain globally ERF, it weakens some local attentions. By contrast, our proposed LKM-UNet with large kernel Mamba achieves larger ERF both in global and local aspects.

\section{Conclusions}
In this paper, we introduced a new Mamba-based UNet model for medical image segmentation, achieving large kernel spatial modeling. Further, we designed a bidirectional and hierarchical SSM to enhance the capacities of Mamba in local and global feature modeling.  Comprehensive experiments on multi-organ segmentation datasets demonstrated the effectiveness of our proposed method. 

\begin{credits}
\subsubsection{\ackname} This research was partially supported by National Natural Science Foundation of China under grants No. 62176231, No. 62106218, No. 82202984, No. 92259202 and No. 62132017, Zhejiang Key R\&D Program of China under grant No. 2023C03053.\footnote{The authors have no competing interests to declare that are
relevant to the content of this article.}
\end{credits}

\bibliographystyle{splncs04}
\bibliography{refs}
\end{document}

\title{Supplemental Document for Submission \#286: \\``LKM-UNet: Large Kernel Vision Mamba UNet for Medical Image Segmentation''}

\author{Jinhong Wang\inst{1,2,5} \and
Jintai Chen\inst{3}\textsuperscript{(\Letter)} \and
Danny Chen\inst{4} \and
Jian Wu\inst{5,6}
\thanks{\Letter : Corresponding Author.}
}

\authorrunning{J. Wang et al.}

\institute{College of Computer Science and Technology, Zhejiang University, China\\ \and 
Institute of Wenzhou, Zhejiang University, China \\ \and
Computer Science Department, University of Illinois Urbana-Champaign, USA\\ \and
Department of Computer Science and Engineering, University of Notre Dame, USA \\ \and
State Key Laboratory of Transvascular Implantation Devices of The Second Affiliated Hospital, Zhejiang University School of Medicine, China\\ \and
School of Public Health Zhejiang University, China\\ 
\email{wangjinhong@zju.edu.cn}, \email{jtchen721@gmail.com}, \email{dchen@nd.edu}, \email{wujian2000@zju.edu.cn}}

\maketitle              % typeset the header of the contribution

\section{More Experiments on Abdomen CT Dataset.}
Here, we show more kernel size analysis and ablation study results on the Abdomen CT Dataset to further demonstrate the effectiveness of our method and its components.

\subsection{Kernel Size Analysis on Abdomen CT Dataset.}
As the same as the main article, we explore \texttt{LKM-UNet}'s performance in different kernel size settings on on Abdomen CT Dataset, whose stage number is 6. Table~\ref{tab3} shows the results. Comparing the performances of the three kernel-size settings, a similar conclusion can be found that \texttt{LKM-UNet} with larger kernel sizes achieves better performances. This indicates that large receptive fields are critical for both 2d and 3d medical image segmentation.

\renewcommand{\cmidrulesep}{0.5mm} %定义两条相邻
\setlength{\aboverulesep}{0.5mm} %在线条[不包括
\setlength{\belowrulesep}{0.5mm} %在线条[不包括

% \usepackage{multirow}
\begin{table}[]
\centering
\caption{Performances of \texttt{LKM-UNet} in three different kernel size settings. The kernel size sequence indicates the kernel size in each stage.}
\label{tab3}
\tabcolsep=0.2cm
\resizebox{0.9\textwidth}{!}{%
\begin{tabular}{c|c|c|c}
\toprule
\multirow{2}{*}{Kernel size} &
  \multirow{2}{*}{\begin{tabular}[c]{@{}c@{}}(5, 7, 6), (5, 7, 6), \\ (5, 7, 6), (5, 7, 6), \\ (5, 7, 6), (5, 7, 6)]\end{tabular}} &
  \multirow{2}{*}{\begin{tabular}[c]{@{}c@{}}[(10, 14, 12), (10, 14, 12), \\ (10, 14, 12), (5, 7, 6), \\ (5, 7, 6), (5, 7, 6)]\end{tabular}} &
  \multirow{2}{*}{\begin{tabular}[c]{@{}c@{}}[(20, 28, 24), (20, 28, 24), \\ (10, 14, 12), (10, 14, 12), \\ (5, 7, 6), (5, 7, 6)]\end{tabular}} \\
  &       &   &   \\
    &       &     &       \\ \midrule
DSC & 86.18 & 86.45  & {\bf 86.82} \\ \midrule
NSD & 89.75   & 89.89  & {\bf 90.02} \\ \bottomrule
\end{tabular}
}
\end{table}

\begin{figure}[t]
\centering
\includegraphics[width=0.98\textwidth]{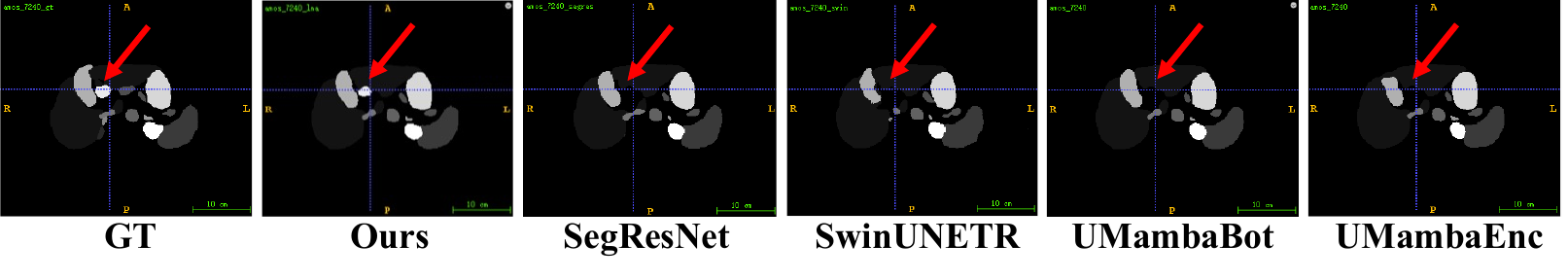}
\caption{Qualitative segmentation visualization of previous methods and our LKM-UNet.}
\label{vis22}
\end{figure}

\subsection{Ablation Study on Abdomen CT Dataset.}
We also conduct ablation experiments on the Abdomen CT dataset to show the importance of each component of \texttt{LKM-UNet}. Table~\ref{tab2} shows the results. We can also find that both PiM and PaM provide improvements for \texttt{LKM-UNet} over the baseline model, while PiM gains more improvements than PaM, suggesting that enlarging the receptive field of local feature modeling is a key to improving model performance. After introducing BiM, the performance of \texttt{LKM-UNet} further improves, which shows the importance of bidirectional Mamba for location-aware sequence modeling. Finally, \texttt{LKM-UNet} with all the components also achieves the best performance, which further demonstrates the effectiveness of our method and its components.

% \usepackage{multirow}
\begin{table}[t]
\centering
\caption{Performances of \texttt{LKM-UNet} with different sub-modules. PiM = Pixel-level SSM. PaM = Patch-level SSM. BiM = Bidirectional Mamba.}
\label{tab2}
\tabcolsep=0.2cm
\resizebox{\textwidth}{!}{%
\begin{tabular}{c|c|c|c|cl|c|c|c}
\toprule
\multirow{2}{*}{Method} &
  \multirow{2}{*}{Baseline} &
  \multirow{2}{*}{Only PiM} &
  \multirow{2}{*}{Only PaM} &
  \multicolumn{2}{c|}{\multirow{2}{*}{PiM + BiM}} &
  \multirow{2}{*}{PaM + BiM} &
  \multirow{2}{*}{PiM + PaM} &
  \multirow{2}{*}{PiM + PaM + BiM} \\
    &       &   &   & \multicolumn{2}{c|}{}     &     &  &   \\ \midrule
DSC & 86.15 & 86.54 & 86.42 & \multicolumn{2}{c|}{86.70} & 86.60 & 86.73 & {\bf 86.82} \\ \midrule
NSD & 89.72     & 89.85 & 89.78 & \multicolumn{2}{c|}{89.99} & 89.95 & 90.00 & {\bf 90.02} \\ \bottomrule
\end{tabular}
}
\vskip -0.8 em
\end{table}

\section{Qualitative Segmentation Results}
To show a more detailed performance of segmentation, we show the qualitative segmentation 
visualization of previous methods and our LKM-UNet in Fig.~\ref{vis22}. It can be seen that our LKM-UNet can recognize the small organ and segment it well which shows LKM-UNet is stronger in not only global modeling but also local details, further demonstrating the effectiveness of LKM-UNet by its large kernel Mamba design.

% \usepackage{multirow}